\documentclass[11pt]{article}

\usepackage[T1]{fontenc}
\usepackage[utf8]{inputenc}
\usepackage{lmodern}
\usepackage[margin=1in]{geometry}
\usepackage{amsmath,amssymb}
\usepackage{graphicx}
\usepackage{booktabs}
\usepackage{makecell}
\usepackage{array}
\usepackage[table]{xcolor}
\usepackage{authblk}
\usepackage[labelfont=bf,labelsep=period,font=small,justification=justified,singlelinecheck=false]{caption}
\usepackage{float}
\usepackage{cite}
\usepackage{enumitem}
\usepackage{microtype}
\usepackage[hidelinks]{hyperref}

\definecolor{diagblue}{HTML}{DAEEF3}
\definecolor{overorange}{HTML}{FDE9D9}
\definecolor{underred}{HTML}{F2DBDB}

\newcommand{\kap}{$\kappa$}

\hypersetup{
  pdftitle={Multi-institutional grading of Barrett’s esophagus dysplasia without stain normalization using a pathology foundation model and multiple instance learning},
  pdfauthor={Amirali Khosravi et al.}
}

\title{\textbf{Multi-institutional grading of Barrett’s esophagus dysplasia without stain normalization using a pathology foundation model and multiple instance learning}}

\author[1]{Amirali Khosravi, MD}
\author[1]{Jaidip M. Jagtap, PhD}
\author[2]{D. Chamil Codipilly, MD}
\author[3]{Catherine E. Hagen, MD}
\author[4]{Jason T. Lewis, MD}
\author[5]{Rickey E. Carter, PhD}
\author[1,6]{Shahriar Faghani, MD}
\author[7]{Vani J.A. Konda, MD}
\author[8]{Fouad Otaki, MD}
\author[9]{Shajan Peter, MD}
\author[10]{Nicholas Shaheen, MD, MPH}
\author[11]{Sachin Wani, MD}
\author[12]{Prasad G. Iyer, MD, MSc}
\author[1]{Bradley J. Erickson, MD, PhD}

\affil[1]{Department of Radiology, Mayo Clinic, Rochester, MN}
\affil[2]{Division of Gastroenterology \& Hepatology, Mayo Clinic, Rochester, MN}
\affil[3]{Department of Laboratory Medicine and Pathology, Mayo Clinic, Rochester, MN}
\affil[4]{Department of Laboratory Medicine and Pathology, Mayo Clinic, Jacksonville, FL}
\affil[5]{Department of Quantitative Health Sciences, Mayo Clinic, Jacksonville, FL}
\affil[6]{Department of Radiology, University of Pennsylvania, Philadelphia, PA}
\affil[7]{Department of Gastroenterology, Baylor Scott and White Center for Esophageal Diseases at Baylor University Medical Center, Dallas, TX}
\affil[8]{Division of Gastroenterology and Hepatology, Oregon Health and Science University, Portland, OR}
\affil[9]{Division of Gastroenterology, University of Alabama Birmingham, Birmingham, AL}
\affil[10]{Center for Esophageal Diseases and Swallowing, Division of Gastroenterology and Hepatology, University of North Carolina Chapel Hill, Chapel Hill, NC}
\affil[11]{Division of Gastroenterology and Hepatology, University of Colorado Anschutz Medical Campus, Aurora, CO}
\affil[12]{Barrett's Esophagus Unit, Division of Gastroenterology and Hepatology, Mayo Clinic, Scottsdale, AZ USA}

\date{}

\begin{document}
\maketitle

\vspace{-2em}
\begin{center}
\small
\textbf{Corresponding author:} Prasad G. Iyer MD MSc\\
Professor, Mayo Clinic Alix School of Medicine\\
Chair, Division of Gastroenterology and Hepatology\\
Mayo Clinic, Scottsdale, AZ\\
Email: \href{mailto:iyer.prasad@mayo.edu}{iyer.prasad@mayo.edu}
\end{center}

\begin{abstract}
\noindent Barrett's esophagus dysplasia grading suffers substantial inter-observer variability, particularly for low-grade dysplasia, complicating clinical decision-making. We conducted a retrospective multi-institutional diagnostic accuracy study of a foundation model-based deep learning system for dysplasia grading, using 969 H\&E-stained whole slide images from four institutions. Patch-level classifiers trained on expert annotations identified diagnostically relevant regions, followed by attribute-aware multiple instance learning for slide-level classification as non-dysplastic Barrett's esophagus (NDBE), low-grade dysplasia, or high-grade dysplasia. On a held-out test set (n = 130), macro-averaged AUC was 0.96 (95\% CI 0.93--0.98). For binary dysplasia detection, sensitivity was 98\% (80/82) and specificity 81\% (39/48), with a negative likelihood ratio of 0.030 and quadratic-weighted \kap{} of 0.83 against expert consensus. No high-grade dysplasia case was misclassified as NDBE. Performance was consistent across institutions without stain normalization, indicating that stain-robust foundation-model representations may remove the site-specific preprocessing that has constrained prior systems to institution-specific deployment.
\end{abstract}

\section{Introduction}

Barrett's esophagus (BE) affects approximately 1-3\% of the general population in the United States \cite{r1,r2,r3} and represents the primary precursor lesion to esophageal adenocarcinoma (EAC), with annual progression rates ranging from 0.3\% per year for non-dysplastic Barrett's esophagus to 28.6\% for high-grade dysplasia \cite{r4,r5,r6,r7}. Dysplasia grading determines clinical management: non-dysplastic cases require surveillance every 3-5 years, low-grade dysplasia prompts either annual surveillance or endoscopic eradication therapy, and high-grade dysplasia necessitates endoscopic eradication therapy \cite{r8}. However, substantial inter-observer variability in dysplasia grading complicates clinical decision-making \cite{r9}. Among expert gastrointestinal pathologists, agreement is substantial for high-grade dysplasia (\kap{} = 0.71) and non-dysplastic Barrett's esophagus (\kap{} = 0.79) but falls to fair or poor levels for low-grade dysplasia (\kap{} = 0.42) \cite{r10,r11}. Community pathologists demonstrate even lower concordance with expert consensus, with documented tendency toward dysplasia overdiagnosis \cite{r12}. This variability leads to unnecessary referrals, patient anxiety, and potentially inappropriate interventions.

Recent applications of deep learning to Barrett's esophagus dysplasia classification have shown promising results \cite{r13,r14,r15,r16}. Prior work from our group demonstrated the feasibility of deep learning for BE dysplasia grading \cite{r13}; however, external validation across institutions \cite{r15} required institution-specific stain normalization models to account for differences in tissue processing and scanner protocols, an approach that is impractical for broad deployment across diverse pathology laboratories. The emergence of foundation models in computational pathology addresses this limitation directly, providing stain-robust pre-trained representations that generalize across institutions without site-specific adaptation \cite{r17,r18}. Despite these advances, multi-institutional training approaches combining foundation models with clinical utility evaluation for community practice settings remain underexplored.

We aimed to develop a foundation model-based deep learning system for Barrett's esophagus dysplasia grading using a multi-institutional training and validation dataset. The system employs a two-stage architecture: patch-level classifiers trained on expert annotations identify diagnostically relevant regions, followed by attribute-aware multiple instance learning for slide-level classification \cite{r19}. We evaluated the system's performance across institutions, characterized clinical utility metrics including likelihood ratios for dysplasia detection, and analyzed error patterns to assess safety for potential clinical deployment.

\section{Methods}

\subsection{Dataset \& Preprocessing}

\subsubsection{Ethics declaration}

This study was conducted in accordance with the Declaration of Helsinki and was approved by the Mayo Clinic Institutional Review Board (approval no.\ 23-009891). Because the study involved retrospective analysis of existing pathology data and posed no more than minimal risk, the Mayo Clinic Institutional Review Board waived the requirement for informed consent and for HIPAA authorization.

\subsubsection{Dataset}

This retrospective study utilized a multi-institutional dataset comprising 969 Barrett's esophagus (BE) slides from 961 patients across four centers: Mayo Clinic (n = 531), Baylor Dallas (n = 169), Oregon Health \& Science University (OHSU) (n = 151), and University of Alabama at Birmingham (UAB) (n = 118). BE diagnosis was confirmed through review of endoscopic and histologic reports. Inclusion criteria were: (1) endoscopic evidence of BE ($\geq$1 cm columnar mucosa extending proximally from the gastroesophageal junction), and (2) confirmed histologic diagnosis of BE (intestinal-type mucosa with goblet cells). Individual patients could contribute multiple slides from different specimens obtained during the same or different procedures, but not from multiple levels (serial sections) of a single specimen. The highest grade of dysplasia on each slide, as confirmed by consensus among expert study pathologists, was used for classification. Patient demographics and endoscopic findings by institution are summarized in \textbf{Table~\ref{tab:S1}}.

Two expert pathologists (CH, JL) with significant BE dysplasia experience independently reviewed all slides to confirm dysplasia grade based on established criteria \cite{r27,r28}. Discordant cases were resolved through discussion between reviewers, with the consensus read serving as the final label. Slides were classified as non-dysplastic BE (NDBE), low-grade dysplasia (LGD), or high-grade dysplasia (HGD). Mayo Clinic slides were annotated for regions with the highest dysplasia grade using Aperio ImageScope software (Leica Biosystems), as previously described \cite{r13}. Slides from non-Mayo institutions lacked patch-level annotations and contributed slide-level dysplasia grade labels only.

The dataset was randomly split into training ($\sim$70\%), validation ($\sim$15\%), and test ($\sim$15\%) sets, stratified by pathology grade and institution, and grouped by patient ID to prevent data leakage. The final distribution comprised 709 training slides (268 NDBE, 233 LGD, 208 HGD), 130 validation slides (49 NDBE, 41 LGD, 40 HGD), and 130 test slides (48 NDBE, 45 LGD, 37 HGD).

\subsubsection{Preprocessing pipeline}

Whole slide images scanned at 40$\times$ magnification were preprocessed to extract 1280$\times$1280 pixel tissue patches using a grid-based tiling approach with 50\% overlap between adjacent tiles. This patch size was chosen empirically so that individual glandular tissues of the specimen would fit into a single patch. Tissue segmentation was performed using HSV color space thresholding, where background regions were identified when pixels met either: (1) brightness $>$240 or $<$30 combined with saturation $<$76, or (2) brightness $>$200 with saturation $<$45. Tiles containing $\geq$5\% tissue were retained for analysis to ensure comprehensive tissue inclusion.

\subsection{Feature Extraction}

Features were extracted from each patch using Midnight-12k \cite{r24}, a pathology-specific foundation model. Midnight is a self-supervised vision transformer trained on over 12,000 whole slide images using the DINOv2 framework, demonstrating robust performance in tissue classification and cancer detection tasks without requiring manual annotations.

\subsection{Model Framework}

\begin{figure}[htbp]
\centering
\includegraphics[width=\textwidth]{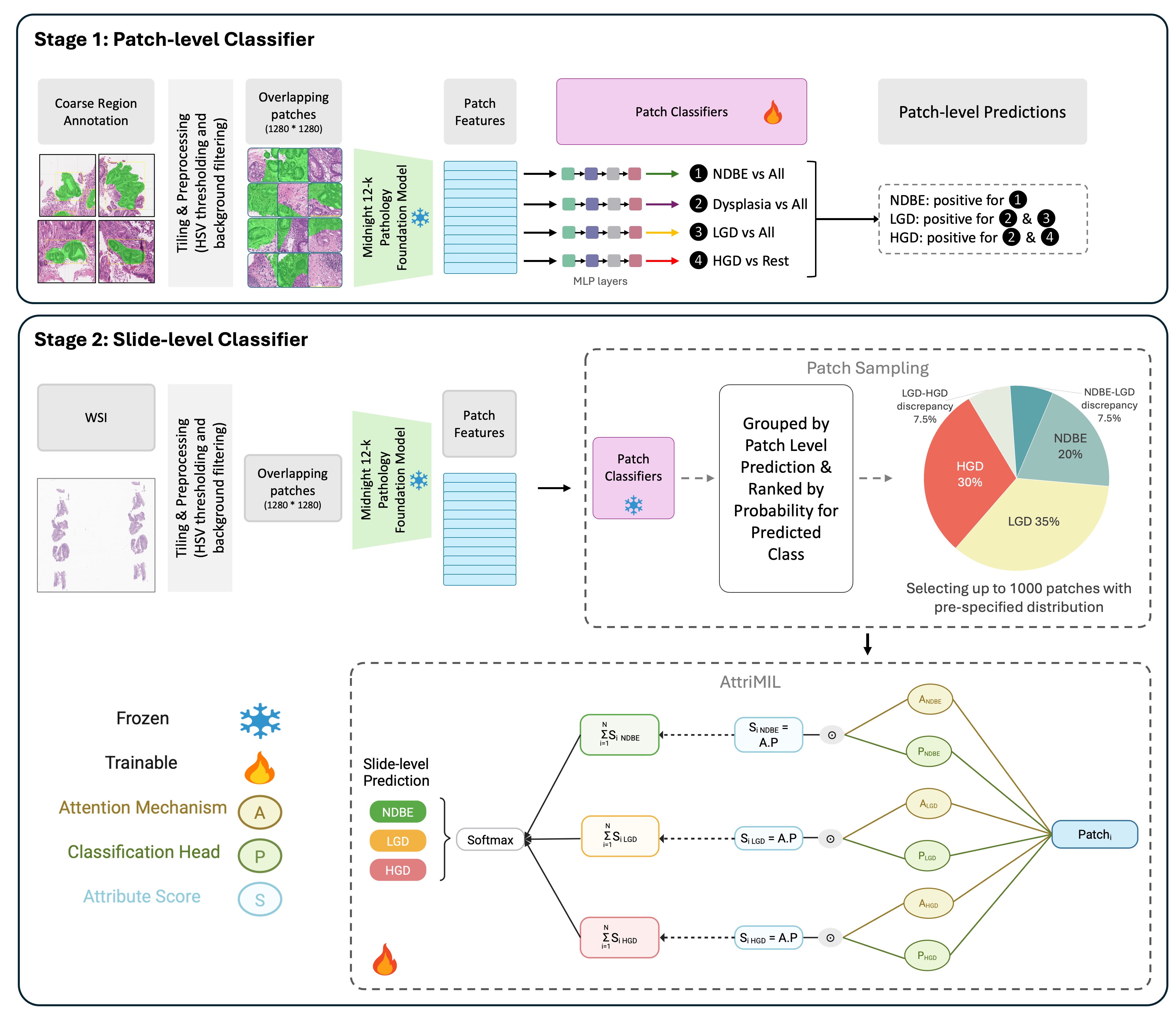}
\caption[Two-stage deep learning framework for Barrett's esophagus dysplasia grading.]{\textbf{Two-stage deep learning framework for Barrett's esophagus dysplasia grading.} \textbf{(a) Stage 1: Patch-level classifier.} Annotated Mayo Clinic slides were tiled into overlapping patches, embedded using the frozen Midnight-12k foundation model, and classified by four MLP classifiers (NDBE, dysplasia, LGD, HGD vs.\ all) to assign each patch a final category. \textbf{(b) Stage 2: Slide-level classifier.} Patches from each whole-slide image were embedded and scored by the frozen Stage 1 classifiers, then sampled into a 1000-patch bag per slide using a pre-specified class distribution. The bag was input to AttriMIL, which computes class-specific attention (A) and instance-level scores (P) to generate attribute scores (S), aggregated and passed through softmax to yield the final slide-level prediction (NDBE, LGD, or HGD).\par
Snowflakes denote frozen components (weights held fixed, not updated during training); flames denote trainable components (weights updated during training)\newline
BE: Barrett's esophagus; NDBE: non-dysplastic BE; LGD: low-grade dysplasia; HGD: high-grade dysplasia; MIL: multiple instance learning; MLP: Multilayer Perceptron}
\label{fig:framework}
\end{figure}

The classification framework consisted of two sequential stages (\textbf{Figure~\ref{fig:framework}}). Stage 1 employed patch-level classifiers to identify diagnostically relevant tissue regions and select informative patches for each slide. Stage 2 utilized a multiple instance learning (MIL) architecture to generate slide-level predictions from the sampled patches.

\subsection{Stage 1: Patch-Level Classification and Sampling}

\subsubsection{Patch-Level Classifier Training}

Stage 1 classifiers were trained using annotated Mayo Clinic slides. Slides contained a median of 3 hand-drawn annotation regions (range: 1--28), totaling 1977 annotation regions across 463 slides (662 NDBE, 753 LGD, 562 HGD). Bounding boxes were fitted around these annotations, and 1280$\times$1280-pixel patches were extracted. Patches were assigned to one of four classes based on tissue content and annotation overlap:

\begin{itemize}[itemsep=0pt]
  \item Background: patches with $\geq$5\% tissue and zero pixels within annotated regions from NDBE slides
  \item NDBE: patches with $\geq$40\% tissue and $\geq$25\% pixels within annotated regions from NDBE slides
  \item LGD: patches with $\geq$40\% tissue and $\geq$25\% pixels within annotated regions from LGD slides
  \item HGD: patches with $\geq$40\% tissue and $\geq$25\% pixels within annotated regions from HGD slides
\end{itemize}

Four binary classifiers were trained using Midnight-extracted features as input:

\begin{enumerate}[label=(\arabic*),itemsep=0pt]
  \item NDBE classifier: NDBE vs non-NDBE (background, LGD, or HGD)
  \item Dysplasia classifier: dysplasia (LGD or HGD) vs non-dysplasia (background or NDBE)
  \item LGD classifier: LGD vs non-LGD (background, NDBE, or HGD)
  \item HGD classifier: HGD vs non-HGD (background, NDBE, or LGD)
\end{enumerate}

Each classifier was a small neural network (a two-layer multilayer perceptron) that learns to weigh combinations of the extracted image features and output a probability for each patch category; standard training techniques (ReLU activation, batch normalization, and dropout) were used to stabilize learning and reduce overfitting. Patches were assigned final classifications based on the four classifier outputs, as summarized in \textbf{Table~\ref{tab:patchrules}}.

The dysplasia classifier was included to reduce false positives in LGD and HGD classifiers, which occasionally misclassified artifacts or high-cellularity non-glandular regions as dysplastic tissue.

\begin{table}[htbp]
\centering
\small
\renewcommand{\arraystretch}{1.3}
\begin{tabular}{@{}lcccc@{}}
\toprule
Assigned Class & \makecell{NDBE\\Classifier} & \makecell{Dysplasia\\Classifier} & \makecell{LGD\\Classifier} & \makecell{HGD\\Classifier} \\
\midrule
\textbf{NDBE} & \boldmath$\checkmark$ & -- & -- & -- \\
\textbf{LGD} & -- & \boldmath$\checkmark$ & \boldmath$\checkmark$ & -- \\
\textbf{HGD} & -- & \boldmath$\checkmark$ & -- & \boldmath$\checkmark$ \\
\textbf{LGD-HGD discrepancy} & -- & \boldmath$\checkmark$ & \boldmath$\checkmark$ & \boldmath$\checkmark$ \\
\textbf{LGD-NDBE discrepancy} & \boldmath$\checkmark$ & \boldmath$\checkmark$ & \boldmath$\checkmark$ & -- \\
\textbf{HGD-NDBE discrepancy} & \boldmath$\checkmark$ & \boldmath$\checkmark$ & -- & \boldmath$\checkmark$ \\
\textbf{Background} & \multicolumn{4}{c}{All other combinations} \\
\bottomrule
\end{tabular}
\caption{\textbf{Patch classification criteria based on binary classifier outputs.} `$\checkmark$' indicates a positive classifier prediction. Discrepancy categories reflect conflicting classifier outputs. NDBE: Non-dysplastic Barrett's Esophagus; LGD: Low-grade Dysplasia; HGD: High-grade Dysplasia}
\label{tab:patchrules}
\end{table}

\subsubsection{Patch Sampling Strategy}

For Stage 2 MIL training, a priority-based sampling strategy selected up to 1000 patches per slide to form a ``bag,'' the fixed-size collection of patches used to represent that slide during classification. Patches were first categorized into six groups: NDBE, LGD, HGD, and three discrepancy categories (LGD-HGD, LGD-NDBE, HGD-NDBE) (\textbf{Table~\ref{tab:patchrules}}). Spatial redundancy was removed by identifying duplicates in 3$\times$3 neighborhoods; when multiple neighboring patches belonged to the same category, only the patch with highest prediction probability was retained. Patches were sorted by prediction probability, with non-duplicate patches prioritized to ensure spatial diversity.

Bags were constructed with the following composition: 20\% NDBE, 35\% LGD, 30\% HGD, 7.5\% LGD-HGD discrepancy, and 7.5\% LGD-NDBE discrepancy. NDBE-HGD discrepancy patches were excluded as they typically represented model errors or artifacts. For each category, candidate pools were created by oversampling at 1.1$\times$ the target number, from which the exact target was randomly drawn to introduce training variability. When slides had insufficient patches in a category, deficits were filled from combined discrepancy categories or neighboring diagnostic categories following a predefined priority order (HGD filled from LGD, then NDBE). Remaining positions were padded with null embeddings randomly selected from background tissue patches to maintain the 1000-patch bag size and improve model robustness to non-tissue regions. Final bags were shuffled before input to prevent positional bias.

\subsection{Stage 2: Multiple Instance Learning (MIL) Architecture}

The Stage 2 model employed AttriMIL \cite{r19}, an MIL framework in which the algorithm learns which patches in a slide's bag are most diagnostically informative for each grade, similar to a pathologist scanning a slide and focusing on the most relevant regions, rather than weighing every patch equally. AttriMIL scores each patch's relevance to NDBE, LGD, and HGD separately (class-specific attention) and combines this with a per-patch prediction (instance classifier) into a combined ``attribute score'' for each class, which is aggregated across the bag to produce the final slide-level diagnosis.

\subsection{Training Details}

Models were trained using the AdamW optimizer with learning rate $1\times10^{-5}$ and weight decay $3\times10^{-4}$. Training proceeded for 400 epochs with early stopping (patience = 40 epochs) using distributed data parallel strategy across four A100 GPUs via PyTorch. The best model checkpoint was selected based on validation macro-average F1 score.

\subsection{Statistical Analysis}

Performance metrics (area under the receiver operating characteristic curve (AUC), sensitivity, specificity, F1 score, and accuracy) were reported with 95\% confidence intervals computed via bootstrap resampling (n = 1000 iterations, random seed=42). Macro-averaged AUC was computed as the unweighted mean of per-class one-versus-rest AUC values.

Agreement between model predictions and expert pathologist consensus with categories for dysplasia grading (NDBE, LGD, and HGD) was assessed using Cohen's \kap{} with unweighted and quadratic weighted schemes, with quadratic weighting selected as the primary measure to appropriately penalize non-adjacent errors in ordinal dysplasia grades. Per-class \kap{} was calculated using one-versus-rest binarization for each grade, with an additional binary \kap{} computed for distinguishing any dysplasia (LGD+HGD) from NDBE. Kappa values were interpreted as none (0.00-0.20), minimal (0.21-0.40), weak (0.41-0.60), moderate (0.61-0.80), or strong (0.81-0.90), and almost perfect agreement (above 0.90) \cite{r29}. Clinical utility was assessed using positive and negative likelihood ratios (LR+/LR$-$) for binary dysplasia grade prediction. Cross-institutional consistency was assessed by comparing per-site accuracy using Fisher's exact test and the predicted probability assigned to the true class using the Kruskal-Wallis test, the latter chosen because these values were not normally distributed and institutional group sizes were small and unequal. All statistical tests were two-tailed with significance threshold $\alpha$ = 0.05.

\subsection{Code availability}

Code for model training and evaluation is available from the corresponding author upon reasonable request. This study used the publicly available Midnight-12k foundation model \cite{r24} and the published AttriMIL architecture \cite{r19}.

\section{Results}

\subsection{Overall Performance}

The test set comprised 130 slides with the following grade distribution: 48 NDBE (36.9\%), 45 LGD (34.6\%), and 37 HGD (28.5\%) from four institutions. Overall performance metrics are presented in \textbf{Table~\ref{tab:performance}}. The model achieved strong discriminative performance across all three classes, with AUC ranging from 0.94 for LGD (the most diagnostically challenging category) to 0.97 for both NDBE and HGD, yielding a macro-averaged AUC of 0.96 (95\% CI: 0.93, 0.98) and macro-averaged F1 of 0.85 (95\% CI: 0.78, 0.91). LGD demonstrated the lowest F1 score (0.82; 95\% CI: 0.72, 0.90), consistent with its known diagnostic difficulty. For binary dysplasia (LGD/HGD vs.\ NDBE) detection, sensitivity was 98\% (80/82; 95\% CI: 0.94, 1.00) and specificity was 81\% (39/48; 95\% CI: 0.70, 0.93). Per-class sensitivity was 89\% (33/37; 95\% CI: 0.78, 0.98) for HGD and 84\% (38/45; 95\% CI: 0.74, 0.94) for LGD. The model achieved a positive LR of 5.20 (95\% CI: 3.23, 13.06) and negative LR of 0.030 (95\% CI: 0.013, 0.078) for any dysplasia detection. At the dysplasia prevalence reported for index endoscopy in Barrett's esophagus, approximately 13\% to 17\% \cite{r20,r21}, a negative model prediction reduces the post-test probability to 0.5\% to 0.6\% and a positive prediction increases it to 44\% to 51\%; after a negative prediction it remains below 1.3\% even at an assumed prevalence of 30\%.

\begin{table}[htbp]
\centering
\small
\begin{tabular}{@{}lcccccc@{}}
\toprule
\textbf{Class} & \textbf{AUC} & \textbf{Sensitivity} & \textbf{Specificity} & \textbf{F1 Score} & \makecell{\textbf{Positive}\\\textbf{Likelihood}\\\textbf{Ratio}} & \makecell{\textbf{Negative}\\\textbf{Likelihood}\\\textbf{Ratio}} \\
\midrule
\textbf{NDBE} & \makecell{0.97\\(0.93, 0.99)} & \makecell{0.81\\(0.70, 0.93)} & \makecell{0.98\\(0.94, 1.00)} & \makecell{0.88\\(0.80, 0.95)} & -- & -- \\ \addlinespace
\textbf{LGD} & \makecell{0.94\\(0.89, 0.98)} & \makecell{0.84\\(0.74, 0.94)} & \makecell{0.88\\(0.81, 0.95)} & \makecell{0.82\\(0.72, 0.90)} & -- & -- \\ \addlinespace
\textbf{HGD} & \makecell{0.97\\(0.94, 0.99)} & \makecell{0.89\\(0.78, 0.98)} & \makecell{0.91\\(0.85, 0.97)} & \makecell{0.85\\(0.76, 0.92)} & -- & -- \\ \addlinespace
\makecell[l]{\textbf{Dysplasia}\\\textbf{(HGD/LGD)}} & \makecell{0.97\\(0.93, 0.99)} & \makecell{0.98\\(0.94, 1.00)} & \makecell{0.81\\(0.70, 0.93)} & \makecell{0.94\\(0.89, 0.97)} & \makecell{5.20\\(3.23, 13.06)} & \makecell{0.03\\(0.01, 0.08)} \\ \addlinespace
\makecell[l]{\textbf{Average}\\\textbf{(Macro)}} & \makecell{0.96\\(0.93, 0.98)} & \makecell{0.85\\(0.78, 0.91)} & \makecell{0.92\\(0.89, 0.95)} & \makecell{0.85\\(0.78, 0.91)} & & \\
\bottomrule
\end{tabular}
\caption{\textbf{Model Performance on Test Set}. Values are reported with 95\% confidence intervals. Overall AUC was 0.96 (95\% CI: 0.93, 0.98).\newline
AUC: area under the receiver operating characteristic curve; NDBE: Non-dysplastic Barrett's Esophagus; LGD: Low-grade Dysplasia; HGD: High-grade Dysplasia.}
\label{tab:performance}
\end{table}

\subsection{Model-Pathologist Agreement}

The model demonstrated moderate to strong agreement with expert pathologist consensus, achieving quadratic weighted Cohen's \kap{} of 0.83 (95\% CI: 0.72, 0.91) and unweighted \kap{} of 0.77 (95\% CI: 0.67, 0.86). Per-class \kap{} values were consistently high: NDBE 0.81 (95\% CI: 0.70, 0.92), LGD 0.72 (95\% CI: 0.58, 0.83), and HGD 0.78 (95\% CI: 0.65, 0.89). For binary dysplasia detection, \kap{} was 0.81 (95\% CI: 0.70, 0.92).

Exact agreement occurred in 84.6\% of cases (110/130), with 85\% of misclassifications (17/20) involving adjacent dysplasia grades (\textbf{Table~\ref{tab:confusion}}). Of the 20 misclassified cases, 14 (70\%) were upgraded and 6 (30\%) were downgraded relative to expert consensus. Only three cases (2.3\%) exhibited non-adjacent errors (NDBE$\leftrightarrow$HGD), and no HGD cases were misclassified as NDBE, reflecting a conservative classification bias that favors over-grading over under-grading.

\begin{table}[htbp]
\centering
\small
\renewcommand{\arraystretch}{1.5}
\begin{tabular}{@{}lcccc@{}}
\toprule
 & \makecell{\textbf{Predicted}\\\textbf{NDBE}} & \makecell{\textbf{Predicted}\\\textbf{LGD}} & \makecell{\textbf{Predicted}\\\textbf{HGD}} & \textbf{Total} \\
\midrule
\textbf{True NDBE} & \cellcolor{diagblue}\textbf{39 (81.2\%)} & \cellcolor{overorange}6 (12.5\%) & \cellcolor{overorange}3 (6.2\%) & 48 (100\%) \\
\textbf{True LGD} & \cellcolor{underred}2 (4.4\%) & \cellcolor{diagblue}\textbf{38 (84.4\%)} & \cellcolor{overorange}5 (11.1\%) & 45 (100\%) \\
\textbf{True HGD} & \cellcolor{underred}0 & \cellcolor{underred}4 (10.8\%) & \cellcolor{diagblue}\textbf{33 (89.2\%)} & 37 (100\%) \\
\bottomrule
\end{tabular}
\caption{\textbf{Confusion Matrix Showing Model Predictions Versus Expert Pathologist Consensus.} The model assigned the same dysplasia grade as expert consensus in 110 of 130 cases (84.6\%), corresponding to a quadratic-weighted \kap{} of 0.83 (95\% CI: 0.72, 0.91) and unweighted \kap{} of 0.77 (95\% CI: 0.67, 0.86). Of the 20 discordant cases, 14 (10.8\% of all cases) were overcalled (orange) and 6 (4.6\%) were undercalled (red) relative to expert consensus. NDBE: Non-dysplastic Barrett's Esophagus; LGD: Low-grade Dysplasia; HGD: High-grade Dysplasia.}
\label{tab:confusion}
\end{table}

\subsection{Cross-Institutional Performance}

Per-site macro-averaged AUC ranged from 0.94 to 0.96 with overlapping 95\% confidence intervals, and all three grades were represented at every institution (\textbf{Table~\ref{tab:sites}}). Accuracy did not differ significantly across sites (Baylor 71\%, Mayo 87\%, OHSU 86\%, UAB 89\%; Fisher's exact p = 0.34), nor did the predicted probability assigned to the true class (Kruskal-Wallis H = 2.82, p = 0.42). The lower F1 at Baylor (0.68; 95\% CI 0.44, 0.88) reflects the small site sample (n = 21, with only 3 HGD cases), which makes F1 sensitive to individual misclassifications.

\begin{table}[htbp]
\centering
\small
\begin{tabular}{@{}lccccc@{}}
\toprule
\textbf{Institution} & \makecell{\textbf{N}\\\textbf{Slides}} & \makecell{\textbf{Class}\\\textbf{Distribution}\\\textbf{(NDBE/LGD/HGD)}} & \makecell{\textbf{AUC}\\\textbf{(95\% CI)}} & \makecell{\textbf{F1 Score}\\\textbf{(95\% CI)}} & \makecell{\textbf{Accuracy}\\\textbf{(95\% CI)}} \\
\midrule
\textbf{Mayo Clinic} & 68 & 19/30/19 & \makecell{0.96\\(0.93, 0.99)} & \makecell{0.87\\(0.78, 0.94)} & \makecell{0.87\\(0.78, 0.94)} \\ \addlinespace
\textbf{Baylor} & 21 & 11/7/3 & \makecell{0.94\\(0.85, 1.00)} & \makecell{0.68\\(0.44, 0.88)} & \makecell{0.71\\(0.52, 0.90)} \\ \addlinespace
\textbf{OHSU} & 22 & 7/5/10 & \makecell{0.96\\(0.89, 1.00)} & \makecell{0.84\\(0.64, 1.00)} & \makecell{0.86\\(0.73, 1.00)} \\ \addlinespace
\textbf{UAB} & 19 & 11/3/5 & \makecell{0.96\\(0.87, 1.00)} & \makecell{0.86\\(0.60, 1.00)} & \makecell{0.89\\(0.74, 1.00)} \\
\bottomrule
\end{tabular}
\caption{\textbf{Model Performance Across Four Medical Institutions.} Values are reported with 95\% confidence intervals. Accuracy compared across institutions by Fisher's exact test (p = 0.34); predicted probability for the true class compared by Kruskal-Wallis test (p = 0.42).\newline
NDBE: Non-dysplastic Barrett's Esophagus; LGD: Low-grade Dysplasia; HGD: High-grade Dysplasia; OHSU: Oregon Health \& Science University; UAB: University of Alabama at Birmingham.}
\label{tab:sites}
\end{table}

\subsection{Attention Map Visualization}

\begin{figure}[htbp]
\centering
\includegraphics[width=\textwidth]{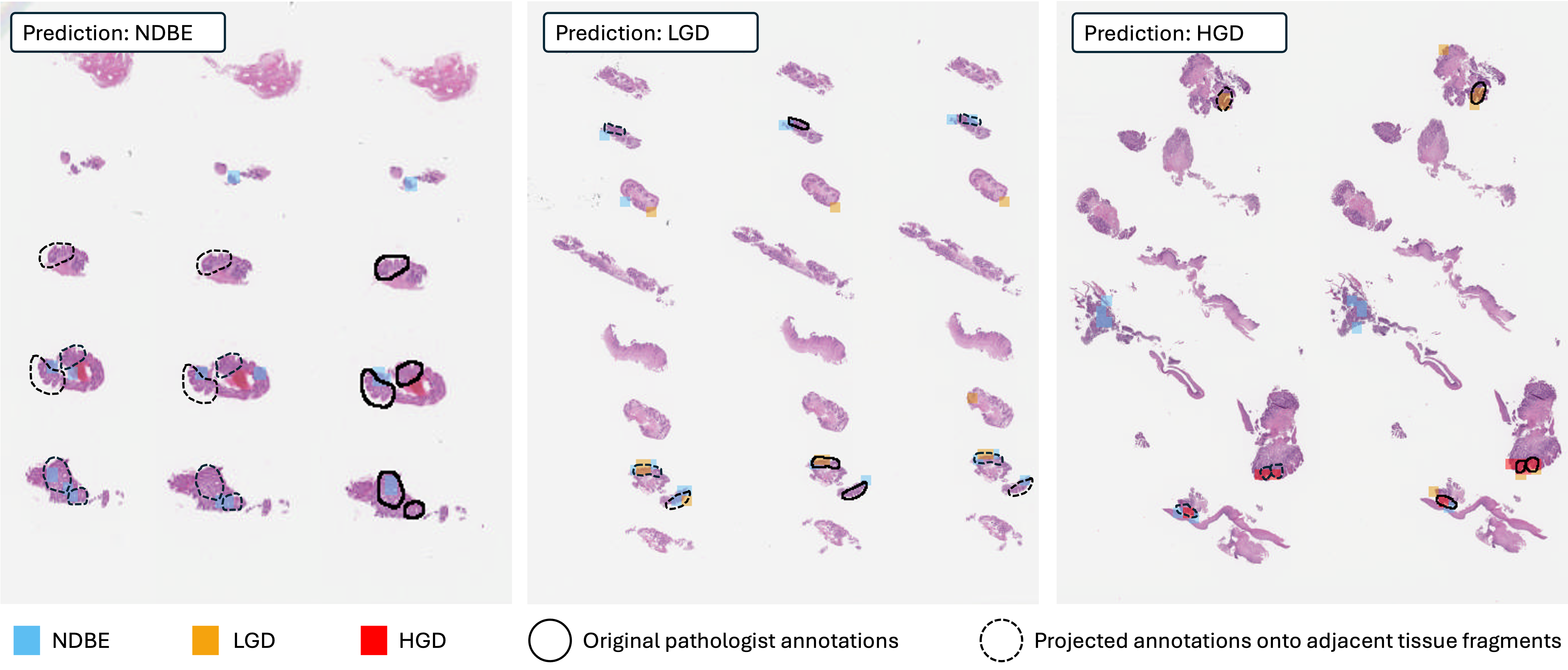}
\caption{\textbf{Attention map visualization for correctly classified cases.} Each column represents a slide with multiple tissue fragments from serial sections. Solid black outlines indicate regions annotated by pathologists as containing the highest dysplasia grade. Dashed black outlines indicate corresponding regions on adjacent tissue fragments. Color overlay shows the top 20 patches with highest attribute scores for each class: blue (NDBE), orange (LGD), red (HGD). High-attention regions demonstrate qualitative overlap with pathologist-annotated dysplastic areas.}
\label{fig:attention}
\end{figure}

Representative attention maps from correctly classified Mayo Clinic test cases are shown in \textbf{Figure~\ref{fig:attention}}. High-attention regions demonstrated qualitative overlap with pathologist-annotated dysplastic areas, suggesting the model attended to diagnostically relevant tissue features. Additional examples of accurate classifications and misclassified cases are provided in Figure~\ref{fig:S1}-\ref{fig:S6}.

\section{Discussion}

This study demonstrates that a foundation model-based approach can grade Barrett's esophagus dysplasia across four institutions without the institution-specific stain normalization required by prior deep learning systems. Using stain-robust pretrained representations, the model achieved a macro-averaged AUC of 0.96 and 98\% sensitivity for dysplasia detection, with a negative likelihood ratio of 0.030 that reduces post-test probability from a reported index-endoscopy dysplasia prevalence of 13\% to 17\% to below 1\% \cite{r20,r21}. Agreement with expert consensus was strong (quadratic-weighted \kap{} = 0.83), performance did not differ significantly across the four contributing institutions (p = 0.34), and the error pattern favored clinical safety: 85\% of misclassifications involved adjacent grades, 70\% were overcalls rather than undercalls, and no high-grade dysplasia case was misclassified as non-dysplastic.

Inter-observer variability in dysplasia grading remains a persistent clinical challenge, with expert agreement particularly poor for low-grade dysplasia (\kap{} = 0.42) \cite{r10} and community pathologists demonstrating lower concordance still, alongside a well-documented tendency toward overdiagnosis \cite{r10,r12}. Difficult cases are typically referred to expert gastrointestinal pathologists for confirmation, but subspecialty expertise is not uniformly available, particularly in community practice. A negative model prediction leaves the probability of dysplasia below 1.5\% across the range of prevalences plausible in Barrett's esophagus populations, low enough to support continued routine surveillance. Because the model assigns a grade rather than a binary label and detected high-grade dysplasia with 89\% sensitivity, cases most likely to require eradication therapy can be prioritized for expert confirmation, which remains necessary because the distinction between low- and high-grade dysplasia determines management. The model's per-class agreement for low-grade dysplasia (\kap{} = 0.72; 95\% CI 0.58, 0.83) compares favorably with published inter-pathologist agreement for this grade (\kap{} = 0.11--0.43) \cite{r10,r11,r22}, although direct comparison requires caution: published values reflect independent reads without adjudication and derive from cohorts with differing case mix, whereas our model was evaluated against adjudicated consensus labels on a class-balanced test set.

The clinical consequence of misclassification depends on its direction. Overcalling to low-grade dysplasia (6 cases) risks unnecessary surveillance intensification or ablation consideration, and overcalling to high-grade dysplasia (8 cases) risks the procedure itself, with real cost, procedural risk, and patient anxiety that should not be minimized. Undercalling high-grade as low-grade dysplasia (4 cases) would substitute surveillance for ablation, potentially delaying treatment and permitting interval progression. Two low-grade dysplasia cases were called non-dysplastic. Because low-grade dysplasia carries lower short-term progression risk than high-grade disease and would in most cases be detected on subsequent surveillance biopsy, the consequence is less severe than a missed high-grade lesion, though not negligible given that low-grade dysplasia is precisely the grade for which the model is intended to add value. These error frequencies reflect the test set class distribution, which does not represent real-world surveillance populations where non-dysplastic Barrett's esophagus predominates; the absolute frequency of overcalls would likely be higher in clinical practice. The complete absence of high-grade-to-non-dysplastic misclassification remains the most important safety finding, as no patient with advanced neoplasia would have been falsely reassured and placed on an extended surveillance interval. While overtreatment carries real costs, missed high-grade dysplasia risks irreversible progression to cancer, making unnecessary intervention the more acceptable failure mode in this setting.

Our group previously developed a two-stage system combining YOLO object detection with ResNet101 classification, achieving greater than 90\% sensitivity and specificity on 542 Mayo Clinic slides \cite{r13}. External validation across three academic centers \cite{r15} confirmed generalizability but required cycle-GAN stain normalization to accommodate inter-institutional staining differences, a step that limits scalability in routine practice. The current study extends this work with a larger multi-institutional training dataset (969 versus 542 slides), a foundation model backbone with built-in stain robustness, removal of the stain normalization requirement, and dysplasia detection sensitivity improving to 98\%. Botros et al.\ developed a system using 290 slides from Amsterdam UMC, achieving AUC 0.94 for distinguishing non-dysplastic Barrett's esophagus from any dysplasia with 92\% sensitivity and 94\% specificity, benchmarked against 55 pathologists \cite{r14}. Our study differs primarily in employing multi-institutional training with a foundation model backbone, whereas Botros et al.\ focused on single-center development with extensive pathologist benchmarking. Collectively these studies establish the feasibility of deep learning for Barrett's esophagus dysplasia grading while highlighting the limitations of single-center development and institution-specific preprocessing \cite{r17,r23}.

Midnight-12k is trained with hue-saturation-density augmentation, exposing the encoder to staining variation during pretraining, and reaches benchmark performance comparable to larger foundation models despite far less pretraining data \cite{r24}. Augmentation of this kind is common across current pathology foundation models \cite{r18}, so these results speak to the class of stain-robust representations rather than to this backbone specifically. The two-stage design balances annotation efficiency against interpretability: Stage 1 classifiers identify diagnostically relevant tissue before Stage 2 aggregation, reducing false-positive classification of artifacts and non-glandular tissue as dysplasia, while AttriMIL's class-specific attention maps allow pathologists to review which regions drove each prediction, supporting clinical oversight rather than black-box deployment. Requiring patch-level annotations from only a single institution while performing consistently across four centers reduces the expert labeling burden that constrains most pathology artificial intelligence systems, consistent with emerging evidence that foundation model features enable few-shot generalization \cite{r17}. Training data were also markedly unbalanced across sites: Mayo Clinic contributed 531 slides, while the other three centers contributed 118 to 169 each. Performance nonetheless remained consistent across institutions (p = 0.34). A model relying on staining or scanning characteristics specific to the dominant center would be expected to underperform at the three smaller sites, and none did.

Several limitations qualify these findings. The retrospective design may introduce slide selection bias, and the near-balanced class distribution of the dataset does not reflect surveillance populations, which affects the absolute error frequencies reported above and the positive predictive value achievable in practice. We assessed slide-level classification rather than patient-level outcomes, and the clinically decisive question, whether algorithmic grading predicts progression to adenocarcinoma, requires longitudinal follow-up unavailable in this dataset. Head-to-head comparison against other contemporary foundation models such as UNI \cite{r17}, Virchow \cite{r23}, or CONCH \cite{r25} was not performed, and ablation studies isolating the contribution of individual architectural components were not conducted; comparable benchmark performance across foundation model backbones in multiple instance learning pipelines \cite{r26} suggests similar results would be obtainable, but this remains untested here. Geographic restriction to four United States centers limits conclusions about institutions using different tissue processing and staining protocols.

Ongoing companion studies address several of these limitations directly: a prospective evaluation of model performance on 450 community-sourced low-grade dysplasia cases will assess real-world generalizability in the population where overcalling is most prevalent and where grade distribution most closely resembles the general Barrett's esophagus population, and a multi-reader multi-case study examining the effect of artificial-intelligence-assisted reading on pathologist performance is underway. Validation in independent international cohorts and patient-level outcome linkage would further establish clinical validity beyond diagnostic accuracy, while uncertainty quantification providing case-specific confidence scores would enable more nuanced decision support. Extension to other gastrointestinal dysplasia contexts, such as gastric intestinal metaplasia or colorectal dysplasia, would test the broader applicability of this approach. Taken together, these results indicate that foundation model-based deep learning can grade Barrett's esophagus dysplasia consistently across four institutions without stain normalization, offering clinically meaningful rule-out capability with a conservative error profile, and that annotations from a single center suffice to reach this performance, a combination that suggests a practical pathway toward scalable deployment of artificial-intelligence-assisted dysplasia grading.

\section{Declaration of generative AI and AI-assisted technologies in the manuscript preparation process}

During the preparation of this manuscript, the authors used Claude (Anthropic) to assist with editing, restructuring, and improving the readability of author-written text. All content, analyses, and interpretations are the original work of the authors, who reviewed and revised all AI-assisted edits and take full responsibility for the final manuscript.

\section{Acknowledgements}

This study was supported in part by the Department of Defense (grant HT9425-23-1079 to P.G.I. and B.J.E.) and the National Cancer Institute (grant 2CA241164 to P.G.I.). The funders played no role in the study design, data collection, analysis and interpretation of data, or the writing of this manuscript.

\section{Author contributions}

\textbf{A.K.:} Conceptualization, Methodology, Software, Formal analysis, Writing -- Original Draft. \textbf{J.M.J.:} Data Curation. \textbf{S.F.:} Conceptualization. \textbf{D.C.C.:} Supervision, Writing -- Review \& Editing. \textbf{C.H., J.L.:} Investigation. \textbf{R.E.C.:} Supervision (statistical analysis), Writing -- Review \& Editing. \textbf{V.J.A.K.:} Resources (Baylor University Medical Center). \textbf{F.O.:} Resources (Oregon Health \& Science University). \textbf{S.P.:} Resources (University of Alabama at Birmingham). \textbf{N.J.S., S.W.:} Writing -- Review \& Editing. \textbf{P.G.I.:} Conceptualization, Supervision, Project administration, Funding acquisition, Writing -- Review \& Editing. \textbf{B.J.E.:} Supervision, Funding acquisition, Writing -- Review \& Editing.

All authors read and approved the final manuscript.

\section{Competing interests}

\begin{itemize}[itemsep=0pt]
  \item SW: Grant support -- Lucid, Phathom Pharmaceuticals, CDx Diagnostics, Exact Sciences, Cyted. Consultant -- Phathom Pharmaceuticals, Boston Scientific, Castle Biosciences, Cook Medical, Lucid, Enterotrack
  \item NJS -- Grant support --Lucid, Interpace Diagnostics, Phathom Pharmaceuticals, Medtronic, Steris, Merit, GIE and CDx Medical. Consultant - Exact Sciences, Phathom Pharmaceuticals, Cook Medical.
  \item VJAK: Consulting : Medtronic, Braintree, Castle, Exact, Ambu
  \item PGI : research funding from Exact Sciences, Pentax Medical, CDx Medical. Consulting with Exact Sciences, Pentax Medical, Merit Medical, CDx Medical and Castle Biosciences
\end{itemize}

All other authors declare no financial or non-financial competing interests.

\renewcommand{\refname}{References}

\clearpage
\setcounter{table}{0}
\setcounter{figure}{0}
\renewcommand{\thetable}{S\arabic{table}}
\renewcommand{\thefigure}{S\arabic{figure}}
\renewcommand{\theHtable}{S\arabic{table}}
\renewcommand{\theHfigure}{S\arabic{figure}}

\section{Supplementary Information}

\begin{table}[H]
\centering
\small
\renewcommand{\arraystretch}{1.3}
\begin{tabular}{@{}lcccc@{}}
\toprule
 & Mayo Clinic & Baylor & OHSU & UAB \\
\midrule
N & 531 & 169 & 151 & 118 \\
Age, mean (SD) & 62.4 (13.7) & 66.6 (11.9) & 68.4 (9.4) & 63.9 (11.8) \\
Male sex, \% & 71.4 \% & 90.5 \% & 87.4 \% & 73.7 \% \\
BE length, cm, mean (SD) & 5.45 (4.0) & 9.2 (5.9) & 7.0 (3.9) & 5.6 (4.2) \\
\bottomrule
\end{tabular}
\caption[Patient Demographics and Endoscopic Findings by Institution.]{Patient Demographics and Endoscopic Findings by Institution.\par
Values are presented as mean (standard deviation) for continuous variables and percentage male for gender.\newline
BE, Barrett's esophagus; OHSU, Oregon Health \& Science University; UAB, University of Alabama at Birmingham.}
\label{tab:S1}
\end{table}

\subsection{Supplementary Figures}

\subsubsection{Note on interpretation of attention overlays (Figures S1--S6)}

Pathologist annotations shown in these figures indicate regions confirmed to contain tissue of the slide's final diagnostic grade. They were not intended as exhaustive delineations of all such tissue on the slide: annotators marked a representative subset of qualifying regions (median 3 per slide, range 1--28), and additional regions of the same grade may be present but unmarked. Each whole slide image also contains several tissue sections cut at different levels from the same specimen, so a lesion annotated on one section is frequently present on adjacent sections without having been separately annotated.

Attention overlays should therefore be read asymmetrically. Overlap between high-scoring patches and annotated regions is positive evidence that the model attended to diagnostically relevant tissue. High-scoring patches falling outside annotated regions do not by themselves indicate that the model attended to irrelevant tissue, since those areas may contain unannotated tissue of the same grade, or the corresponding region of an adjacent section.

\begin{figure}[p]
\centering
\includegraphics[width=\textwidth]{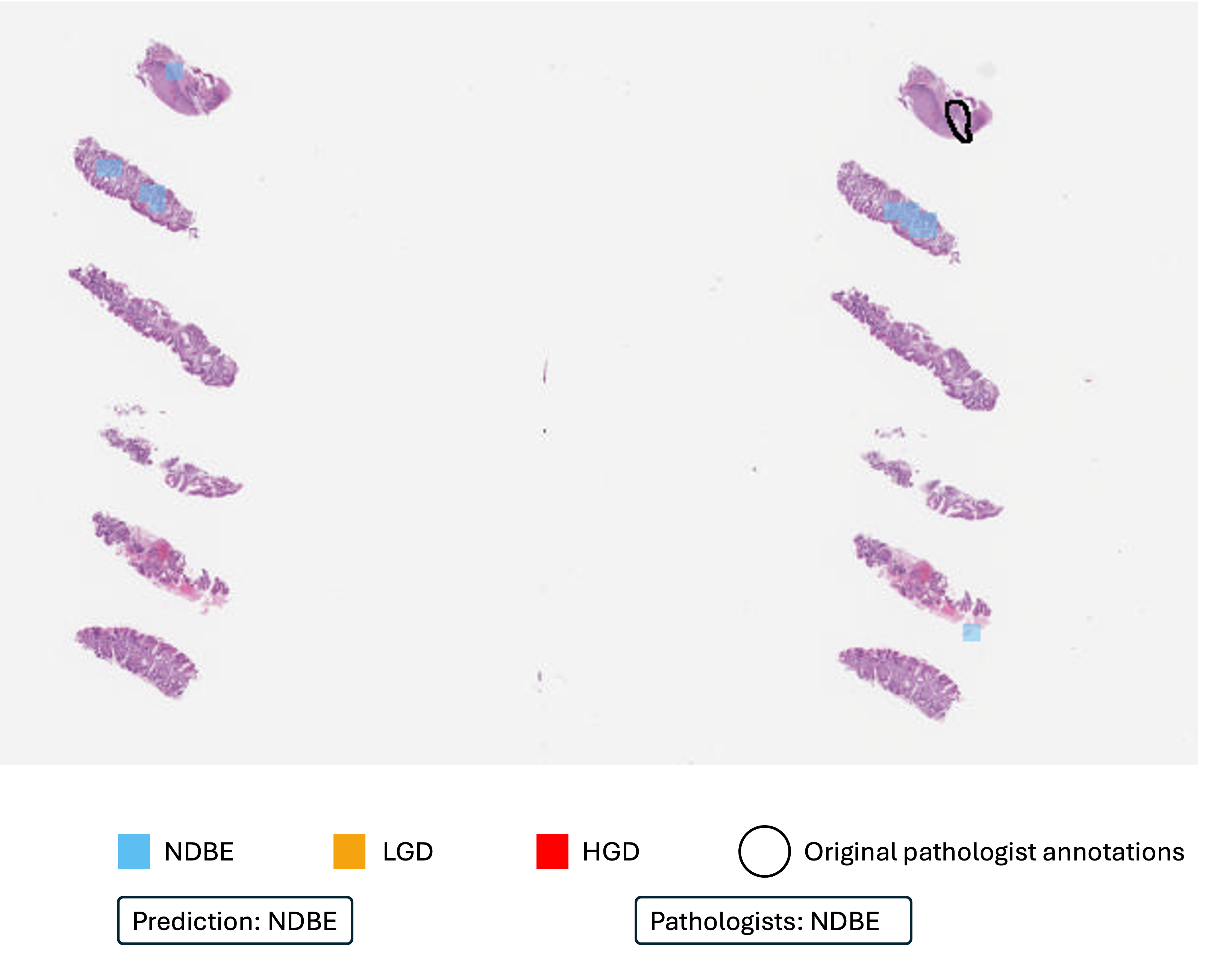}
\caption{Correctly classified non-dysplastic Barrett's esophagus. The model assigned NDBE, in agreement with expert consensus.}
\label{fig:S1}
\end{figure}

\begin{figure}[p]
\centering
\includegraphics[width=\textwidth]{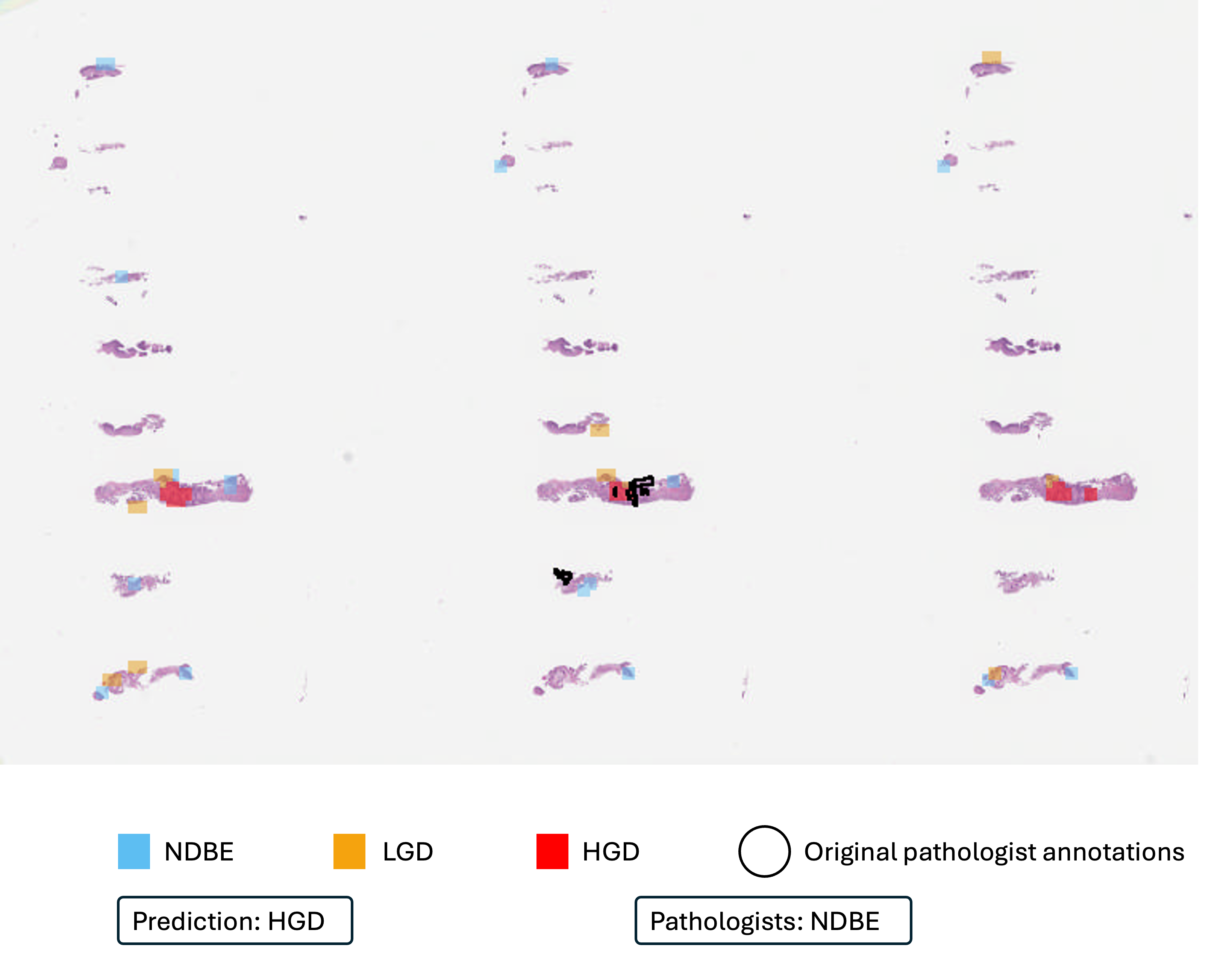}
\caption{Misclassified non-dysplastic Barrett's esophagus. Expert consensus was NDBE; the model predicted HGD.}
\label{fig:S2}
\end{figure}

\begin{figure}[p]
\centering
\includegraphics[width=\textwidth]{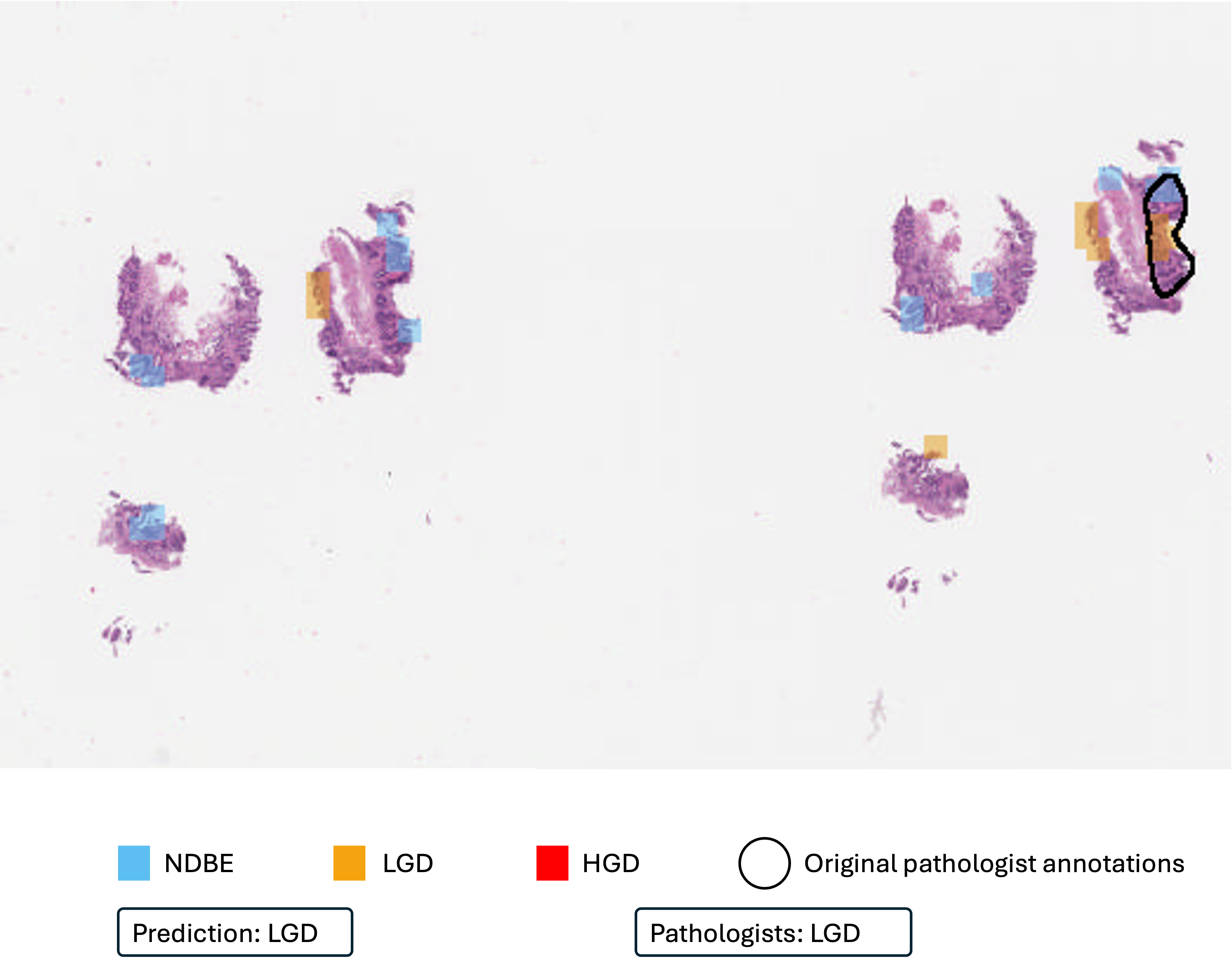}
\caption{Correctly classified LGD. The model assigned LGD, in agreement with expert consensus.}
\label{fig:S3}
\end{figure}

\begin{figure}[p]
\centering
\includegraphics[width=\textwidth]{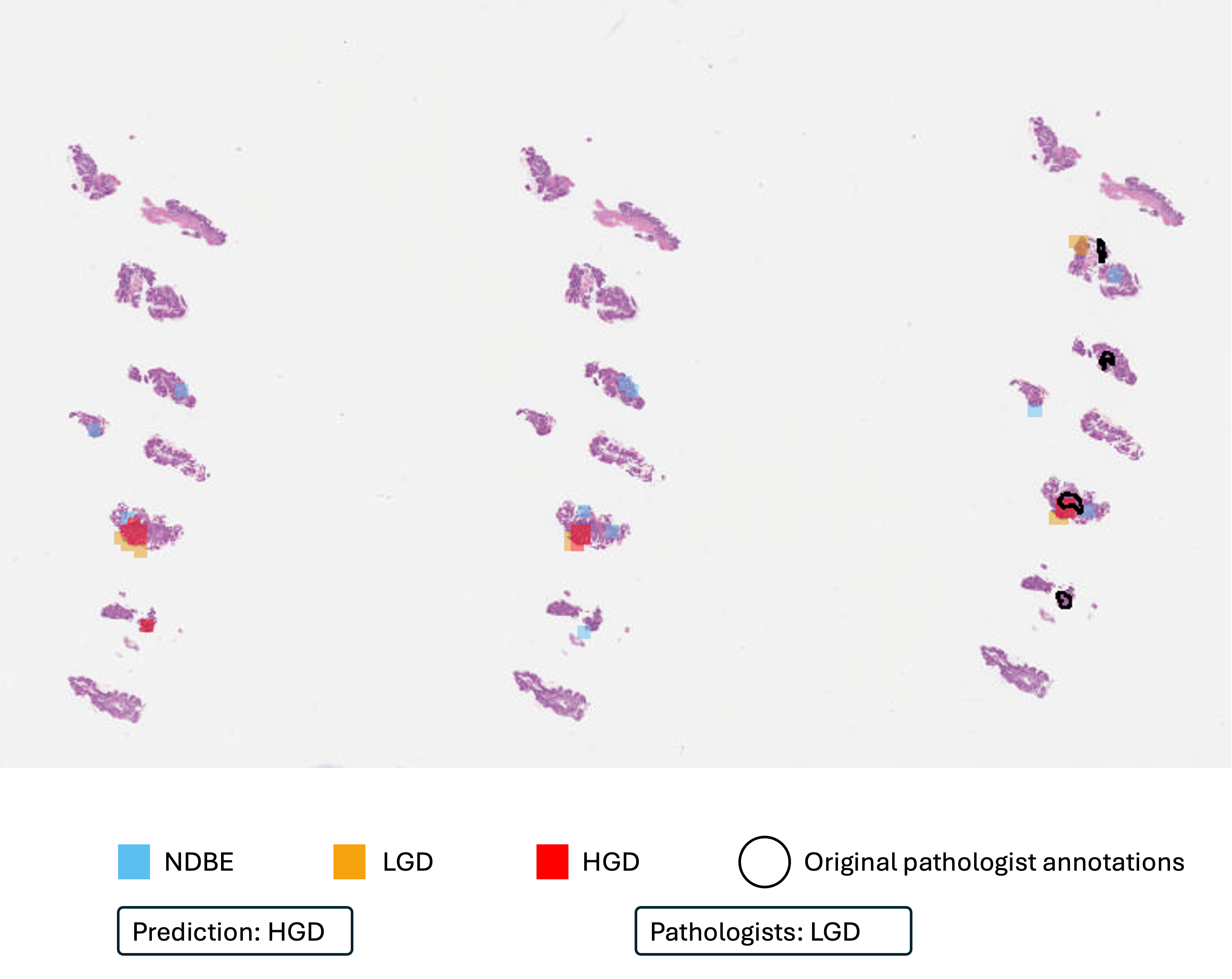}
\caption{Misclassified LGD. Expert consensus was LGD; the model predicted HGD.}
\label{fig:S4}
\end{figure}

\begin{figure}[p]
\centering
\includegraphics[width=\textwidth]{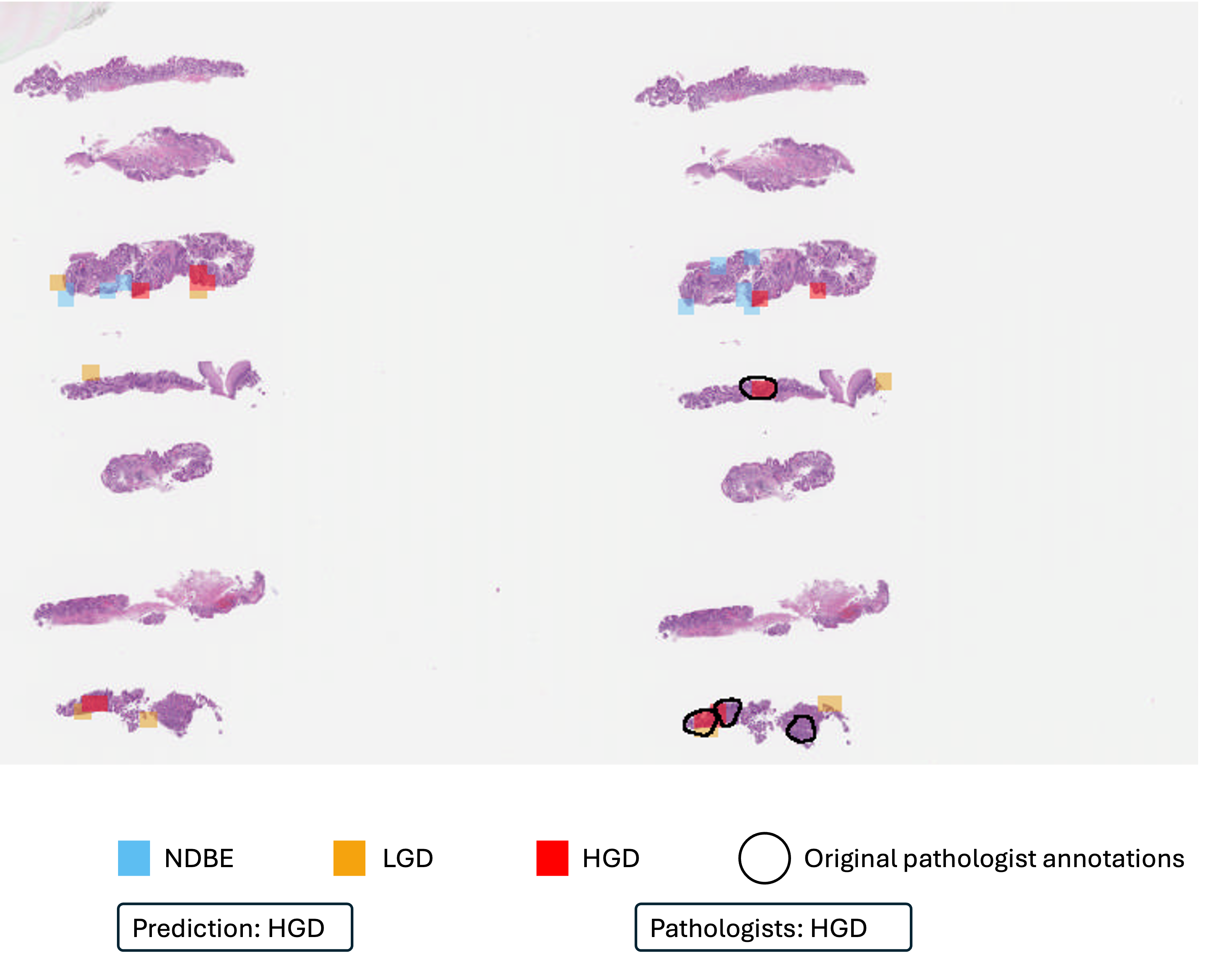}
\caption{Correctly classified HGD. The model assigned HGD, in agreement with expert consensus.}
\label{fig:S5}
\end{figure}

\begin{figure}[p]
\centering
\includegraphics[width=\textwidth]{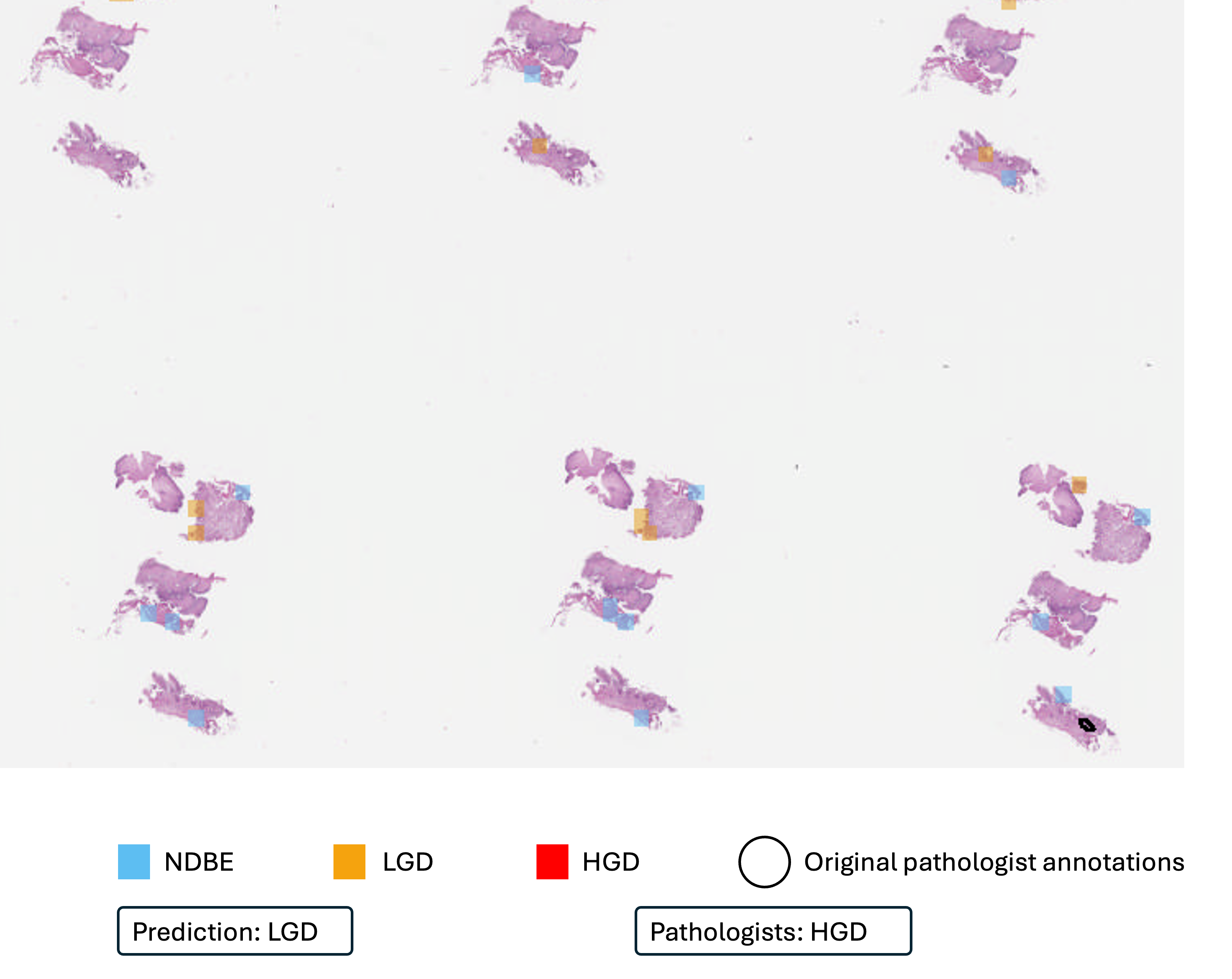}
\caption{Misclassified HGD. Expert consensus was HGD; the model predicted LGD.}
\label{fig:S6}
\end{figure}

\end{document}